%% file: main.tex
\documentclass[10pt,twocolumn,letterpaper]{article}

\usepackage{wacv}
\usepackage{times}
\usepackage{epsfig}
\usepackage{graphicx}
\usepackage{amsmath}
\usepackage{amssymb}

\usepackage{graphicx}
\usepackage{color}
\usepackage{bm}

\usepackage{algorithm}

\usepackage{amsfonts}
\usepackage{multirow}
\usepackage{booktabs}
\usepackage[dvipsnames]{xcolor}
\usepackage{xpatch}
\usepackage{xcolor,colortbl}
\usepackage{array, boldline, makecell}
\usepackage{arydshln}

\def\wacvPaperID{1135} 

\wacvfinalcopy 

\ifwacvfinal
\def\assignedStartPage{1} 
\fi

\ifwacvfinal
\usepackage[breaklinks=true,bookmarks=false]{hyperref}
\else
\usepackage[pagebackref=true,breaklinks=true,colorlinks,bookmarks=false]{hyperref}
\fi

\ifwacvfinal
\else
\fi

\begin{document}

\title{Weakly Supervised Instance Segmentation by Deep Community Learning}

\author{Jaedong Hwang$^1$\thanks{ ~Equal contribution.} \qquad Seohyun Kim$^1$\footnotemark[1]  \qquad Jeany Son$^2$ \qquad Bohyung Han$^1$\\
$^1$ECE \& ASRI, Seoul National University, Seoul, Korea \\
$^2$ETRI, Daejeon, Korea\\
{\tt\small $^1$\{jd730, goodbye61, bhhan\}@snu.ac.kr, $^2$jeany@etri.re.kr}
}

\maketitle

\begin{abstract}
\input{sections/0_abstract}
\end{abstract}

\input{sections/1_introduction.tex}
\input{sections/2_related.tex}

\input{sections/3_method.tex}

\input{sections/4_exp.tex}
\input{sections/5_conclusion.tex}

\vspace{-0.2cm}
\section*{Acknowledgements}
\vspace{-0.2cm}
This work was supported by Naver Labs and Institute for Information \& Communications Technology Promotion (IITP) grant funded by the Korea government (MSIT) [2017-0-01779, 2017-0-01780].

{\small
\bibliographystyle{ieee_fullname}
\bibliography{egbib}
}
\clearpage
\input{sections/6_supp.tex}

\end{document}

%% file: sections/0_abstract.tex

We present a weakly supervised instance segmentation algorithm based on deep community learning with multiple tasks.
This task is formulated as a combination of weakly supervised object detection and semantic segmentation, where individual objects of the same class are identified and segmented separately.
We address this problem by designing a unified deep neural network architecture, which has a positive feedback loop of object detection with bounding box regression, instance mask generation, instance segmentation, and feature extraction.
Each component of the network makes active interactions with others to improve accuracy, and the end-to-end trainability of our model makes our results more robust and reproducible.
The proposed algorithm achieves state-of-the-art performance in the weakly supervised setting without any additional training such as Fast R-CNN and Mask R-CNN on the standard benchmark dataset.
The implementation of our algorithm is available on the project webpage: {{\color{magenta}\url{https://cv.snu.ac.kr/research/WSIS_CL}}}.

%% file: sections/1_introduction.tex
\vspace{-0.2cm}
\section{Introduction}\label{sec:intro}
Object detection and semantic segmentation algorithms have achieved great success in recent years thanks to the advent of large-scale datasets~\cite{everingham15pascal,lin2014microsoft} as well as the development of deep learning technologies~\cite{girshick2015fast,he2017maskrcnn,redmon2018yolo,ren2015faster}.
However, most of existing image datasets have relatively simple forms of annotations such as image-level class labels, while many practical tasks require more sophisticated information such as bounding boxes and areas corresponding to object instances. 
Unfortunately, the acquisition of the complex labels needs significant human efforts, and it is challenging to construct a large-scale dataset containing such comprehensive annotations.

Instead of standard supervised learning formulations~\cite{chen2018masklab,dai2016instance,hayder2017boundary,he2017maskrcnn}, we tackle a more challenging task, weakly supervised instance segmentation, which relies only on image-level class labels for instance-wise segmentation.
This task shares critical limitations with many weakly supervised object recognition problems; trained models typically focus too much on discriminative parts of objects in the scene, and, consequently, fail to identify whole object regions and extract accurate object boundaries in a scene.
Moreover, there are additional challenges in handling two problems jointly, weakly supervised object detection and semantic segmentation; incomplete ground-truths incur noisy estimation of labels in both tasks, which makes it difficult to take advantage of the joint learning formulation.
For example, although object detection methods typically employ object proposals to provide rough information of object location and size, a na\"ive application of instance segmentation module to weakly supervised object detection results may not be successful in practice due to noise in object proposals.

\begin{figure}[t]
\begin{center}
\includegraphics[width=1\linewidth]{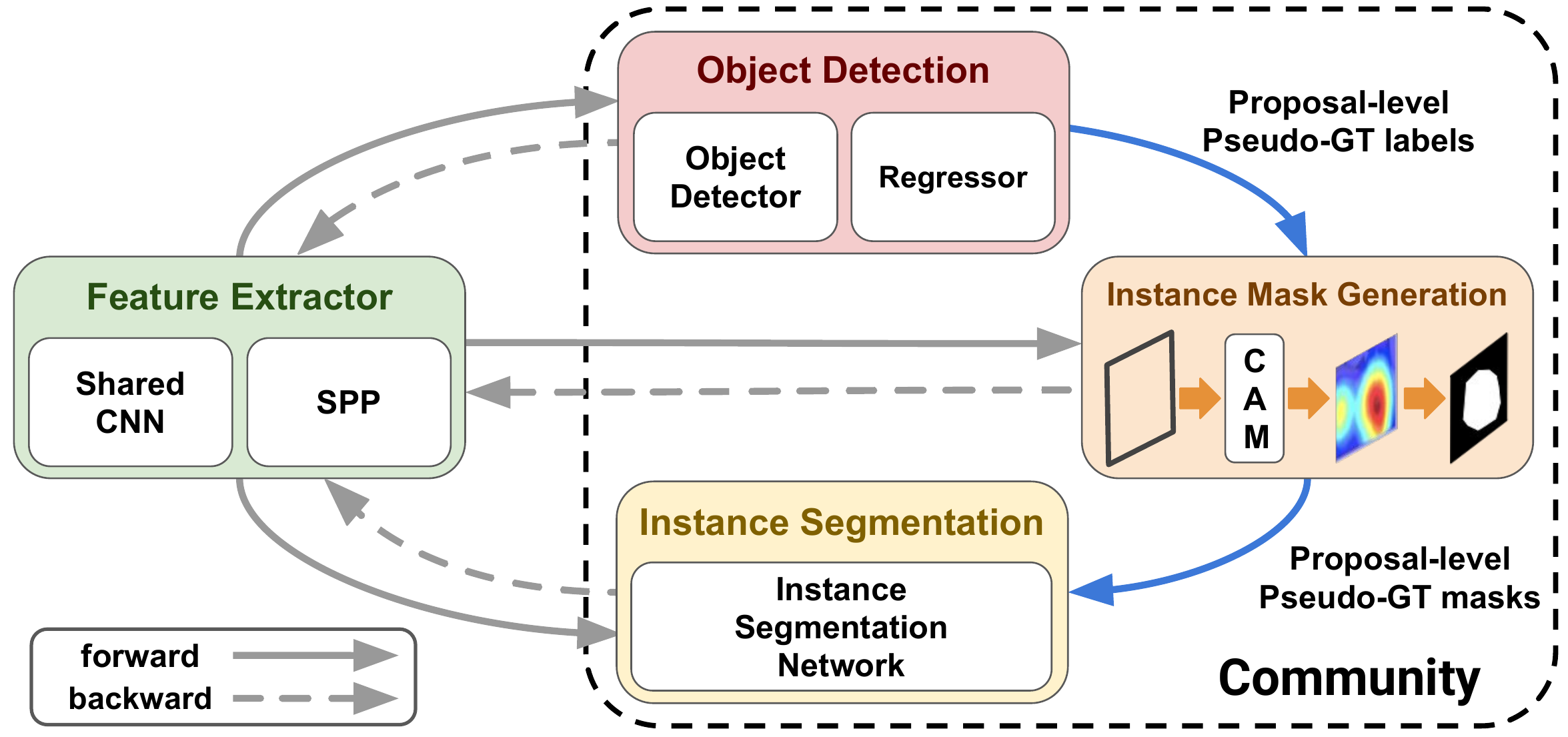}
\end{center}
\vspace{-0.2cm}
   \caption{The proposed community learning framework for weakly supervised instance segmentation. 
   Our model is composed of object detection module, instance mask generation module, instance segmentation module and feature extractor, which constructs a positive feedback loop within a community.
   It first identifies positive detection bounding boxes from the detection module and generates pseudo-ground-truths of instance segmentation using class activation maps. 
   The model is trained with multi-task loss of the three components using the pseudo-ground-truths.
    The final segmentation masks are obtained from the ensemble of outputs from instance mask generation and segmentation modules.
}
\label{fig:concept}
\vspace{-0.3cm}
\end{figure}

Our approach aims to realize the goal using a deep neural network with multiple interacting task-specific components that construct a positive feedback loop.
The whole model is trained in an end-to-end manner and boosts performance of individual modules, leading to outstanding segmentation accuracy.
We call such a learning concept {\it community learning}, and Figure~\ref{fig:concept} illustrates its application to weakly supervised instance segmentation.
The community learning is different from multi-task learning that attempts to achieve multiple objectives in parallel without tight interaction between participating modules.
The contributions of our work are summarized below:
\begin{itemize}
    \item[$\bullet$] We introduce a deep community learning framework for weakly supervised instance segmentation, which is based on an end-to-end trainable deep neural network with active interactions between multiple tasks: object detection, instance mask generation, and object segmentation.
 \item[$\bullet$] We incorporate two empirically useful techniques for object localization, class-agnostic bounding box regression and segmentation proposal generation, which are performed without full supervision.
    \item[$\bullet$]  The proposed algorithm achieves substantially higher performance than the existing weakly supervised approaches on the standard benchmark dataset without post-processing.
\end{itemize}

The rest of the paper is organized as follows.
We briefly review related works in Section~\ref{sec:related} and describe our algorithm with community learning in Section~\ref{sec:method}.
Section~\ref{sec:exp} analyzes the experimental results on a benchmark dataset.

%% file: sections/2_related.tex

\section{Related Works}\label{sec:related}

This section reviews existing weakly supervised algorithms for object detection, semantic segmentation, and instance segmentation.

\subsection{Weakly Supervised Object Detection} 
Weakly Supervised Object Detection (WSOD) aims to localize objects in a scene only with image-level class labels.
Most of existing methods formulate WSOD as Multiple Instance Learning (MIL) problems~\cite{dietterich1997solving} and attempt to learn detection models via extracting pseudo-ground-truth labels~\cite{bilen2016weakly,tang2018pcl,tang2017multiple,zhang2018w2f}.
WSDDN~\cite{bilen2016weakly} combines classification and localization tasks to identify object classes and their locations in an input image.
However, this technique is designed to find only a single object class and instance conceptually and often fails to solve the problems involving multiple labels and objects.
Various approaches~\cite{kantorov2016contextlocnet,son2018forget,tang2018pcl,tang2017multiple,wan2018min} tackle this issue by incorporating additional components, but they are still prone to focus on the discriminative parts of objects instead of whole object regions.
Recently, there are several research integrating semantic segmentation to improve detection performance~\cite{diba2017weakly,li2019weakly,shen2019cyclic,wei2018ts2c}.
WCCN~\cite{diba2017weakly} and TS2C~\cite{wei2018ts2c} filter out object proposals using semantic segmentation results, but still have trouble in identifying spatially overlapped objects in the same class.
Meanwhile, SDCN~\cite{li2019weakly} utilizes semantic segmentation result to refine pseudo-ground-truths.
WS-JDS~\cite{shen2019cyclic} leverages weakly supervised semantic segmentation module that estimates importance for object proposals.
Although the core idea is valuable and the segmentation module improves detection performance, the instance segmentation performance improvement is marginal compared to simple box masking of its baselines~\cite{bilen2016weakly,kantorov2016contextlocnet}.

\subsection{Weakly Supervised Semantic Segmentation}
Weakly Supervised Semantic Segmentation (WSSS) is a task to estimate pixel-level semantic labels in an image based on image-level class labels only.
Class Activation Map (CAM)~\cite{zhou2016learning} is widely used for WSSS because it generates class-specific likelihood maps using the supervision for image classification. 
SPN~\cite{kwak2017weakly}, one of the early works that exploit CAM for WSSS, combines CAM with superpixel segmentation result to extract accurate class boundaries in an image.
AffinityNet~\cite{ahn2018learning} propagates the estimated class labels using semantic affinities between adjacent pixels.
Ge~\etal~\cite{ge2018multi} employ a pretrained object detector to obtain segmentation labels.
Recent approaches~\cite{huang2018weakly,kwak2017weakly,lee2019ficklenet,lee2019frame,wang2018weakly,zeng2019joint} often train their models end-to-end.
DSRG~\cite{huang2018weakly} and MCOF~\cite{wang2018weakly} propose iterative refinement procedures starting from CAM.
FickleNet~\cite{lee2019ficklenet} performs stochastic feature selection in its convolutional layers and captures the regularized shapes of objects.

\subsection{Instance Segmentation}
Instance segmentation can be regarded as a combination of object localization and semantic segmentation, which needs to identify individual object instances.
There exist several fully supervised approaches~\cite{chen2018masklab,dai2016instance,hayder2017boundary,he2017maskrcnn}.
Haydr~\etal~\cite{hayder2017boundary} utilize Region Proposal Network (RPN)~\cite{ren2015faster} to detect individual instances  and leverage Object Mask Network (OMN) for segmentation.
Mask R-CNN~\cite{he2017maskrcnn}, Masklab~\cite{chen2018masklab} and MNC~\cite{dai2016instance} have similar procedures to predict their pixel-level segmentation labels.

There have been recent works for Weakly Supervised Instance Segmentation (WSIS) based on image-level class labels only~\cite{ahn2019weakly,ge2019label,issam2019where,zhou2018weakly,zhu2019learning}.
Peak Response Map (PRM)~\cite{zhou2018weakly} takes the peaks of an activation map as pivots for individual instances and estimates the segmentation mask of each object using the pivots.
Instance Activation Map (IAM)~\cite{zhu2019learning} selects pseudo-ground-truths out of precomputed segment proposals based on PRM to learn segmentation networks.
Label-PEnet~\cite{ge2019label} combines various components with different functionalities to obtain the final segmentation masks. 
However, it involves many duplicate operations across the components and requires very complex training pipeline.
There are a few attempts to generate pseudo-ground-truth segmentation maps based on weak supervision and forward them to the well-established network~\cite{he2017maskrcnn} for instance segmentation~\cite{ahn2019weakly,issam2019where}.
To improve accuracy, the algorithms often employ post-processing such as MCG proposals~\cite{arbelaez2014multiscale} or denseCRF~\cite{krahenbuhl2011efficient}.

%% file: sections/3_method.tex

\section{Proposed Algorithm}\label{sec:method}
This section describes our community learning framework based on an end-to-end trainable deep neural network for weakly supervised instance segmentation.

\subsection{Overview and Motivation}
\label{sub:network}
One of the most critical limitations in a na\"ive combination of detection and segmentation networks for weakly supervised instance segmentation is that the learned models often attend to small discriminative regions of objects and fail to recover missing parts of target objects.
This is partly because segmentation networks rely on noisy detection results without proper interactions and the benefit of the iterative label refinement procedure is often saturated in the early stage due to the strong correlation between outputs from two modules.

To alleviate this drawback, we propose a deep neural network architecture that constructs a circular chain along with the components and generates desirable instance detection and segmentation results.
The chain facilitates the interactions along individual modules to extract useful information.
Specifically, the object detector generates proposal-level pseudo-ground-truth labels.
They are used to create pseudo-ground-truth masks for instance segmentation module, which estimates the final segmentation labels of individual proposals using the masks.
These three network components make up a community and collaborate to update the weights of the backbone network for feature extraction, which leads to regularized representations robust to overfitting to poor local optima.

\begin{figure*}[t]
\begin{center}
\includegraphics[width=0.9\linewidth] {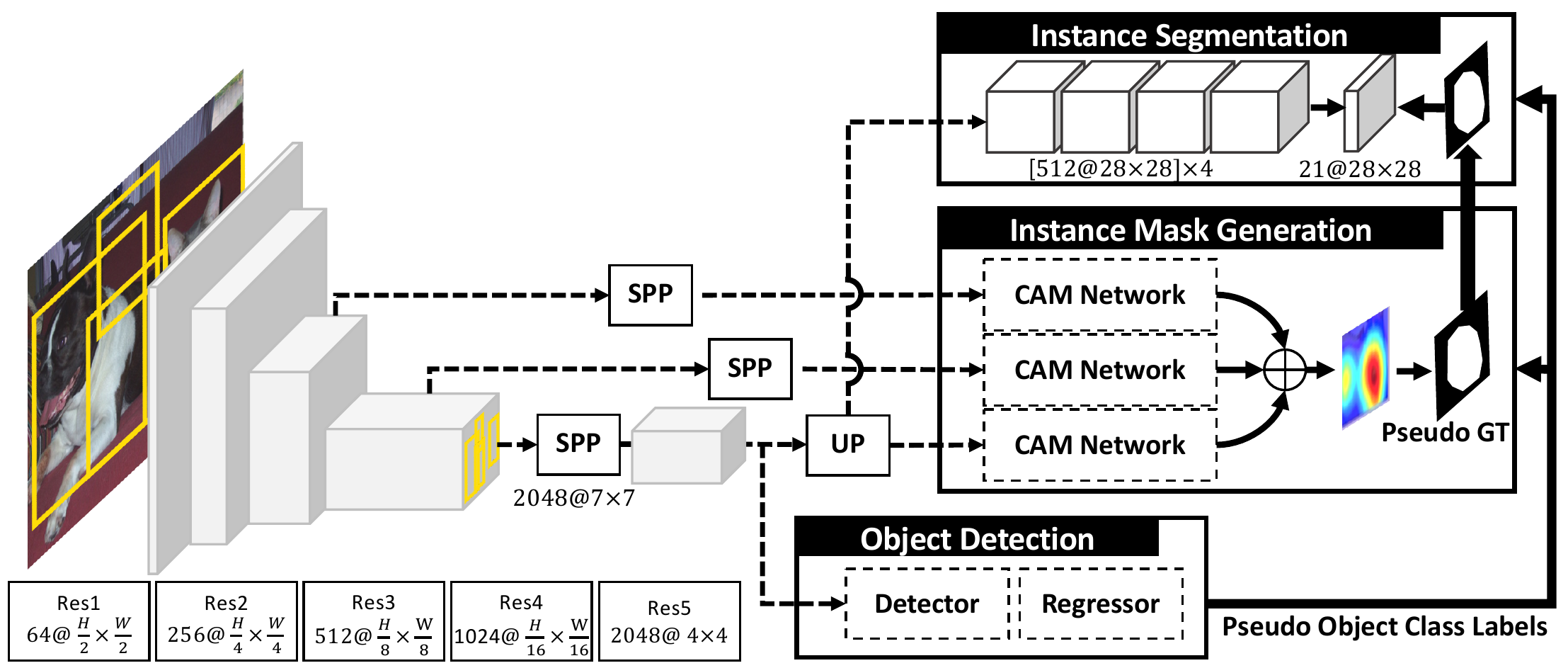}
\end{center}
   \caption{
   The proposed network architecture for weakly supervised instance segmentation. Our end-to-end trainable network consists of four parts:
     (a) \textit{feature extraction network} computes the shared feature maps and provides proposal-level features with the other networks,
     (b) \textit{object detection network} identifies the location of objects and gives a pseudo-label of object class to each proposal,
     (c) \textit{instance mask generation network} constructs the class activation map for given proposals using predicted pseudo-labels from the detector,
     (d) \textit{instance segmentation network} predicts segmentation masks and is learned with the outputs of the above networks as pseudo-ground-truths.
     }
\label{fig:model}
\end{figure*}

\subsection{Network Architecture}
Figure~\ref{fig:model} presents the network architecture of our weakly supervised object detection and segmentation algorithm. 
As mentioned earlier, the proposed network consists of four parts: feature extractor, object detector with bounding box regressor, instance mask generator and instance segmentation module.
Our feature extraction network is made of shared fully convolutional layers, where the feature of each proposal is obtained from the Spatial Pyramid Pooling (SPP)~\cite{he2014spatial} layers on the shared feature map and fed to the other modules.

\paragraph*{3.2.1 Object Detection Module\vspace{0.3cm}\newline}%
\label{par:detection_module} 
For object detection, a $7\times 7$ feature map is extracted from the SPP layer for each object proposal and forwarded to the last residual block (\textsf{res5}).
Then, we pass these features to both the detector and the regressor.
Since this process is compatible with any end-to-end trainable object detection network based on weak supervision, we adopt one of the most popular weakly supervised object detection networks, referred to as OICR~\cite{tang2017multiple}, which has three refinement layers after the base detector.
For each image-level class label, we extract foreground proposals based on their estimated scores corresponding to the label and apply a non-maximum suppression (NMS) to reduce redundancy.
Background proposals are randomly sampled from the proposals that are overlap with foreground proposals below a threshold.
Among the foreground proposals, the one with the highest score for each class is selected as a pseudo-ground-truth bounding box.

Bounding box regression is typically conducted under full supervision to refine the proposals corresponding to objects.
However, learning a regressor in our problem is particularly challenging since it is prone to be biased by discriminative parts of objects; such a characteristic is difficult to control in a weakly supervised setting and is aggravated in class-specific learning.
Hence, unlike~\cite{girshick2015fast,girshick2014rich,ren2015faster}, we propose a class-agnostic bounding box regressor based on pseudo-ground-truths to avoid overly discriminative representation learning and provide better regularization effect. 
Note that a class-agnostic regressor has not been explored actively yet since fully supervised models can exploit accurate bounding box annotations and learning a regressor with weak labels only is not common. 
If a proposal has a higher IoU with its nearest pseudo-ground-truth proposal than a threshold, the proposal and the pseudo-ground-truth proposal are paired to learn the regressor.

\paragraph*{3.2.2 Instance Mask Generation (IMG) Module\vspace{0.3cm}\newline}
\label{par:cam_module}
This module constructs pseudo-ground-truth masks for instance segmentation using the proposal-level class labels given by our object detector.
It takes the feature of each proposal from the SPP layers attached to multiple convolutional layers as shown in Figure~\ref{fig:model}.
Since the IMG module utilizes hierarchical representations from different levels in a backbone network, it can deal with multi-scale objects effectively.

We construct pseudo-ground-truth masks for individual proposals by integrating the following additional features into CAM~\cite{zhou2016learning}.
First, we compute a background class activation map by augmenting a channel corresponding to the background class.
This map is useful to distinguish objects from the background.
Second, instead of the Global Average Pooling (GAP) adopted in the standard CAM, we employ the weighted GAP to give more weights to the center pixels within proposals.
It computes a weighted average of the input feature maps, where the weights are given by an isotropic Gaussian kernel.
Third, we convert input features $f$ of the CAM module to log scale values, \ie, $\log(1+f)$, which penalizes excessively high peaks in the CAM and leads to spatially regularized feature maps appropriate for robust segmentation.

The  output of the IMG module, denoted by $\mathbf{M}$, is an average of three CAMs to which min-max normalizations~\cite{patro2015normalization} are applied. 
For each selected proposal, the pseudo-ground-truth mask  $\mathbf{\widetilde{M}} \in \mathbb{R}^{(C+1) \times T^2}$ for instance segmentation is given by the following equation using the three CAMs, $\mathbf{M}_k$ $(k = 1, 2, 3)$,
\begin{equation}
    \mathbf{\widetilde{M}} = \delta \left[ \frac{1}{3}\sum_{k=1}^3 \mathbf{M}_k > \xi \right],
\label{eq:gt}
\end{equation}
where $\mathbf{M}_k \in \mathbb{R}^{(C+1) \times T^2}$ is the $k^{\text{th}}$ CAM whose size is $T\times T$ for all classes including background, $\delta[\cdot]$ is an element-wise indicator function, and $\xi$ is a predefined threshold.

\paragraph*{3.2.3 Instance Segmentation Module\vspace{0.3cm}\newline}
\label{par:segmentation_module}
For instance segmentation, the output of the \textsf{res5} block is upsampled to $T\times T$ activation maps and provided to four convolutional layers along with ReLU layers and the final segmentation output layer as illustrated in Figure~\ref{fig:model}.
This module learns a pixel-wise binary classification label for each proposal based on the pseudo-ground-truth mask $\mathbf{\widetilde{M}}^c$, provided by the IMG module.
The predicted mask of each proposal is a class-specific binary mask, where the class label $c$ is determined by the detector.
Note that our model is compatible with any semantic segmentation network.

\subsection{Losses}
\label{sub:losses}

The overall loss function of our deep community learning framework is given by the sum of losses from the three modules as
\begin{equation}
\mathcal{L} =  \mathcal{L}_{\text{det}} + \mathcal{L}_{\text{img}} + \mathcal{L}_{\text{seg}},
\label{eq:loss}
\end{equation}
where $\mathcal{L}_{\text{det}}$, $\mathcal{L}_{\text{img}}$, and $\mathcal{L}_{\text{seg}}$  denote detection loss, instance mask generation loss, and instance segmentation loss, respectively.
The three terms interact with each other to train the backbone network including the feature extractor in an end-to-end manner.

\paragraph*{3.3.1 Object Detection Loss\vspace{0.3cm}\newline} 
\label{par:loss_object_detection}
The object detection module is trained with the sum of classification loss $\mathcal{L}_{\text{cls}}$, refinement loss $\mathcal{L}_{\text{refine}}$, and bounding box regression loss $\mathcal{L_{\text{reg}}}$. 
The features extracted from the individual object proposals are given to the detection module based on OICR~\cite{tang2017multiple}.
Image classification loss $\mathcal{L}_\text{cls}$ is calculated by computing the cross-entropy between image-level ground-truth class label $\boldsymbol{y} = (y_1, \dots, y_C)^\text{T}$ and its corresponding prediction $\boldsymbol{\phi} = (\phi_1, \dots, \phi_C)^\text{T}$, which is given by
\begin{equation}
    \mathcal{L}_{\text{cls}} = - \sum_{c=1}^{C} y_c \log \phi_c + (1-y_c)\log(1-\phi_c) \text{,} \\
    \label{eq:Lcls}
\end{equation}
where $C$ is the number of classes in a dataset.
As in the original OICR, the pseudo-ground-truth of each object proposal in the refinement layers is obtained from the outputs of their preceding layers, where the supervision of the first refinement layer is provided by WSDDN~\cite{bilen2016weakly}.
The loss of the $k^\text{th}$ refinement layer is computed by a weighted sum of losses over all proposals as
\begin{equation}
    \mathcal{L}_\text{refine}^k = - \frac{1}{|R|}\sum_{r=1}^{|R|}\sum_{c=1}^{C+1}w_r^k y_{cr}^k \log x_{cr}^{k}, 
    \label{eq:Loicr}
\end{equation}
where $x_{cr}^{k}$ denotes a score of the $r^\text{th}$ proposal with respect to class $c$ in the $k^\text{th}$ refinement layer, $w_r^k$ is a proposal weight obtained from the prediction score in the preceding refinement layer, and $|R|$ is the number of proposals.
In the refinement loss function, there are $C+1$ classes because we also consider background class.

Regression loss $\mathcal{L_{\text{reg}}} $ employs smooth $\ell_1$-norm between a proposal and its matching pseudo-ground-truth, following the bounding box regression literature~\cite{girshick2015fast,ren2015faster}.
The regression loss is defined as follows:
\begin{equation}
\mathcal{L_{\text{reg}}} =\frac{1}{|R|} \sum_{r=1}^{|R|} \sum_{j=1}^{|\mathcal{G}|}  q_{rj} \sum_{k \in \{x,y,w,h\}}  \text{smooth}_{\ell_1}(t_{rjk} - v_{rk}),
\end{equation}
where $\mathcal{G}$ is a set of pseudo-ground-truths, $q_{rj}$ is an indicator variable denoting whether the $r^\text{th}$ proposal is matched with the $j^\text{th}$ pseudo-ground-truth, $v_{rk}$ is a predicted bounding box regression offset of the $k^\text{th}$ coordinate for $r^\text{th}$ proposal and $t_{rjk}$ is the desirable offset parameter of the $k^\text{th}$ coordinate between the $r^\text{th}$ proposal and the $j^\text{th}$ pseudo-ground-truth as in R-CNN~\cite{girshick2014rich}.

The detection loss $\mathcal{L}_\text{det}$ is the sum of image classification loss, bounding box regression loss, and $K$ refinement losses, which is given by 
\begin{equation}
    \mathcal{L}_\text{det} = \mathcal{L}_\text{cls}+ \mathcal{L}_\text{reg} + \sum_{k=1}^K \mathcal{L}_\text{refine}^k,
    \label{eq:Ldet} 
\end{equation}
where $K=3$ in our implementation.

\paragraph*{3.3.2 Instance Mask Generation Loss\vspace{0.3cm}\newline}
\label{par:loss_cam}
For training CAMs in the IMG module, we adopt average classification scores from three refinement branches of our detection network.
The loss function of the $k^\text{th}$ CAM network, denoted by $\mathcal{L}_{\text{cam}}^k$, is given by a binary cross entropy loss as
\begin{align}
    \mathcal{L}_\text{cam}^k = -\frac{1}{|R|} \sum_{r=1}^{|R|}\sum_{c=1}^{C+1} \widetilde{y}_{rc}
    \log p_{rc}^k  + (1-\widetilde{y}_{rc})\log(1-p_{rc}^k) \text{,}
    \label{eq:Lcam_k}
\end{align}
where $\widetilde{y}_{rc}$ is an one-hot encoded pseudo-label from detection branch of the $r^{\text{th}}$ proposal for class $c$, and $p_{rc}^k$ is a softmax score of the same proposal for the same class obtained by the weighted GAP from the last convolutional layer.
The instance mask generation loss is the sum of all the CAM losses as shown in the following equation:
\begin{equation}
    \mathcal{L}_{\text{img}} =  \sum_{k=1}^3 \mathcal{L}_{\text{cam}}^k \text{.}\\
    \label{eq:Lcam}
\end{equation}

\paragraph*{3.3.3 Instance Segmentation Loss\vspace{0.3cm}\newline}
\label{par:loss_instance_segmentation}
The loss in the segmentation network is obtained by comparing the network outputs with the pseudo-ground-truth $\mathbf{\widetilde{M}}$ using a pixel-wise binary cross entropy loss for each class, which is given by 
\begin{align}
    \mathcal{L}_{\text{seg}} =  -\frac{1}{T^2}\sum_r^{|R|}\sum_c^{C+1} \hspace{-0.2cm} &\sum_{(i,j) \in T \times T} \hspace{-0.3cm}  m_{rc}^{ij} \log s_{rc}^{ij}  \\
    &+ (1-m_{rc}^{ij})\log\left(1 - s_{rc}^{ij} \right), \nonumber
\label{eq:Lseg}
\end{align}
where $m_{rc}^{ij}$ means a binary element at $(i,j)$ of $\mathbf{\widetilde{M}}$ for the $r^\text{th}$ proposal, and $s_{rc}^{ij}$ is the output value of the segmentation network, $\mathbf{S} \in \mathbb{R}^{|R| \times (C+1) \times T^2}$, at location $(i,j)$ of the $r^\text{th}$ proposal.

\begin{table*}[t!]
        \caption{Instance segmentation results on the PASCAL VOC 2012 segmentation \textit{val} set with two different types of supervisions ($\mathcal{I}$: image-level class label, $\mathcal{C}$: object count).
        The numbers in red and blue denote the best and the second best scores without Mask R-CNN re-training, respectively.} 
    \label{tab:inst}
        \begin{center}
    \scalebox{0.85}{
    \renewcommand{\arraystretch}{1.1}
   \setlength\tabcolsep{12pt}
   \begin{tabular}{ccc|ccccc}
    \toprule
    Method & Supervision & Post-procesing &  $\text{mAP}_{0.25}$ & $\text{mAP}_{0.5}$ & $\text{mAP}_{0.75}$ & ABO  \\ \hline\hline                       WISE~\cite{issam2019where} w/ Mask R-CNN & $\mathcal{I}$  & \checkmark & 49.2 & 41.7 & 23.7 & 55.2  \\
                 IRN~\cite{ahn2019weakly} w/ Mask R-CNN & $\mathcal{I}$  & \checkmark & - & 46.7 & - & -\\ \hline
    Cholakkal \etal~\cite{cholakkal2019object} &$\mathcal{I} + \mathcal{C}$ &\checkmark & 48.5 & 30.2 & 14.4 &44.3 \\\hdashline
        PRM~\cite{zhou2018weakly} & $\mathcal{I}$& \checkmark & 44.3 & 26.8 & 9.0 & 37.6 \\
    IAM~\cite{zhu2019learning} & $\mathcal{I}$  &\checkmark  & 45.9 & 28.3 & 11.9 & 41.9 \\ 
    Label-PEnet~\cite{ge2019label} & $\mathcal{I}$ &\checkmark & 49.2 & 30.2 & {\color{red}\textbf{12.9}} & 41.4 \\ \hdashline
     \multirow{2}{*}{Ours}& $\mathcal{I}$ && {\color{red}\textbf{57.0}} & {\color{blue}\textbf{35.9}}& 5.8 & {\color{blue}\textbf{43.8}} \\
     & $\mathcal{I}$ & \checkmark &  {\color{blue}\textbf{56.6}} & {\color{red}\textbf{38.1}}& {\color{blue}\textbf{12.3}} & {\color{red}\textbf{48.2}} \\\Xhline{0.6pt}
    \end{tabular}
    }
     \end{center} 
    \vspace{-0.2cm}
\end{table*}

\subsection{Inference}
\label{sub:inference}
Our model sequentially predicts object detection and instance segmentation for each proposal in a given image. 
For object detection, we use the average scores of three refinement branches in the object detection module. 
Each regressed proposal is labeled as the class that corresponds to the maximum score. 
We apply a non-maximum suppression with IoU threshold 0.3 to the proposals.
The survived proposals are regarded as detected objects and used to estimate pseudo-labels for instance segmentation.

For instance segmentation, we select the foreground activation map of the identified class $c$, $\mathbf{M}^c$, from the IMG module and the corresponding segmentation score map, $\mathbf{S}^c$, from instance segmentation module for each detected object.
The final instance segmentation label is given by the ensemble of two results,
\begin{equation}
    \mathbf{O}^c = {\delta} \left[  \frac{\mathbf{M}^c + \mathbf{S}^\text{c}}{2}  >  \xi \right],
    \label{eq:inf}
\end{equation}
where $\mathbf{O}^c$ is a binary segmentation mask for detected class $c$, $\delta[\cdot]$ is an element-wise indicator function, and $\xi$ is a threshold identical used in Eq.~\eqref{eq:gt}.
For post-processing, we utilize Multiscale Combinatorial Grouping (MCG) proposals~\cite{arbelaez2014multiscale} as used in PRM~\cite{zhou2018weakly}. 
 Each instance segmentation mask is substituted as a max overlap MCG proposal.
Since the MCG proposal is a group of superpixels, it contains boundary information.
Hence, if a segmentation output covers overall shape well, MCG proposal is greatly helpful to catch details of an object.

%% file: sections/4_exp.tex
\section{Experiments}\label{sec:exp}

This section describes our setting for training and evaluation and presents the experimental results of our algorithm in comparison to the existing methods. 
We also analyze various aspects of the proposed network.

\subsection{Training}
\label{sub:training}

We use Selective Search~\cite{uijlings2013selective} for generating bounding box proposals.
All fully connected layers in the detection and the IMG modules are initialized randomly using a Gaussian distribution $(0, 0.01^2)$.
The learning rate is $0.001$ at the beginning and reduced to $0.0001$ after 90K iterations.
The hyper-parameter in the weight decay term is $0.0005$, the batch size is 2, and the total training iteration is 120K.
We use 5 image scales of $\{480, 576, 688, 864, 1000\}$, which are based on the shorter size of an image, for data augmentation and ensemble in training and testing. 
The NMS threshold for selecting foreground proposals is 0.3 and $\xi$ in Eq~(\ref{eq:gt}) is set to $0.4$ following MNC~\cite{dai2016instance}.
For regression, a proposal is associated with a pseudo-ground-truth if the IoU is larger than 0.6.
The output size $T$ of the IMG and instance segmentation modules is 28.
Our model is implemented on PyTorch
and the experiments are conducted on a single NVIDIA  Titan XP GPU.

\subsection{Datasets and Evaluation Metrics}
We use PASCAL VOC 2012 segmentation dataset~\cite{everingham15pascal} to evaluate our algorithm.
The dataset is composed of 1,464, 1,449, and 1,456 images for training, validation, and testing, respectively, for 20 object classes.
We use the standard augmented training set (\textit{trainaug}) with 10,582 images to learn our network, following the prior segmentation research~\cite{ahn2019weakly,chen2018deeplab,cholakkal2019object,ge2019label,bharath2011semantic,zhou2018weakly,zhu2019learning}.
In our weakly supervised learning scenario, we only use image-level class labels to train the model.
Detection and instance segmentation accuracies are measured on PASCAL VOC 2012 segmentation validation (\textit{val}) set.

We employ the standard mean average precision (mAP) to evaluate object detection performance, where a bounding box is regarded as a correct detection if it overlaps with a ground-truth more than a threshold, \textit{i.e.} IoU $> 0.5$.
CorLoc~\cite{deselaers2012weakly} is also used to evaluate the localization accuracy on the \textit{trainaug} dataset. 
For instance segmentation task, we evaluate performance of an algorithm using mAPs at IoU thresholds 0.25, 0.5 and 0.75.
We also use Average Best Overlap (ABO) to present overall instance segmentation performance of our model.

\subsection{Comparison with Other Algorithms} \label{sec:comp}
We compare our algorithm with existing weakly supervised instance segmentation approaches~\cite{cholakkal2019object,ge2019label,zhou2018weakly,zhu2019learning}.
Table~\ref{tab:inst} shows that our algorithm generally outperforms the prior arts even without post-processing. 
Note that our post-processing using MCG proposals~\cite{arbelaez2014multiscale} improves mAP at high thresholds and ABO significantly, and leads to outstanding performance in terms of both mAP and ABO after all.
We believe that such large gaps come from the effective regularization given by our community learning.
The accuracy of our model is not as good as the method given by Mask R-CNN re-training~\cite{ahn2019weakly,issam2019where}, but direct comparison is not fair due to the retraining issue.
Table~\ref{tab:inst_train} illustrates that our model outperforms the methods without re-training on \textit{train} split.

\begin{table}[t]
\begin{center}
\caption{Instance segmentation results on the PASCAL VOC 2012 segmentation \textit{train} set. 
\cite{ahn2019weakly,issam2019where} report results without Mask R-CNN obtained from their original papers.}
\label{tab:inst_train}
		\vspace{0.2cm}
\scalebox{0.85}{
\begin{tabular}{cccc}
   \toprule
	& WISE~\cite{issam2019where} & IRN~\cite{ahn2019weakly}  & Ours\\ \hline\hline
  	mAP$_{0.5}$ & 25.8 & 37.7 & \bf{39.2}  \\
   \bottomrule
\end{tabular}
}
\end{center}
\vspace{-0.3cm}
\end{table}

\begin{table}[t]
\begin{center}
\caption{Contribution of individual components integrated into our algorithm study. The evaluation is performed on PASCAL VOC 2012 segmentation \textit{val} set for mAP and \textit{trainaug} set for CorLoc (* indicates that detection bounding boxes are used as segmentation results as well).}
\label{tab:abl_arch}
\scalebox{0.85}{
\begin{tabular}{cccc}
   \toprule
  \multirow{2}{*}{\large Architecture} & $\begin{array}{c} \text{Instance} \\ \text{Segmentation} \end{array}$ &\multicolumn{2}{c}{$\begin{array}{c} \text{Object} \\ \text{Detection} \end{array}$} \\
  		& mAP$_{0.5}$ & mAP & CorLoc \\
   \hline\hline
   Detector & \ 18.8$^*$ &45.3 & 63.6 \\
   Detector + IMG &32.8 &48.6 & 66.3 \\
      Detector +  IMG + IS & 33.7 & 49.2 & 66.8 \\
   Detector + REG +  IMG + IS& \bf{35.9}&\bf{53.2}& \bf{70.8} \\
   \bottomrule
\end{tabular}
}
\end{center}
\vspace{-0.2cm}
\end{table}

\subsection{Ablation Study} \label{sub:abl}
We discuss the contribution of each component in the network and the effectiveness of our training strategy.
We also compare two different regression strategies---class-agnostic vs. class-specific---using detection scores.
Note that we present the results without post-processing for the ablation study to verify the contribution of each component clearly.

\subsubsection{Network Components}
We analyze the effectiveness of individual modules for instance segmentation and object detection.
For comparisons, we measure mAP$_{0.5}$ for instance segmentation and mAP for object detection on PASCAL VOC 2012 segmentation \textit{val} set while computing CorLoc on the \textit{trainaug} set.
Note that the instance segmentation accuracy of the detection-only model is given by using detected bounding boxes as segmentation masks.
All models are trained on PASCAL VOC 2012 segmentation \textit{trainaug} set. 

Table~\ref{tab:abl_arch} presents that the IMG and Instance Segmentation (IS) modules are particularly helpful to improve accuracy for both tasks.
By adding the two components, our model achieves accuracy gain in detection by 3.9\% and 3.2\% points in terms of mAP and CorLoc, respectively, compared to the baseline detector.
Additionally, bounding box regression (REG) enhances performance by generating better pseudo-ground-truths. 

\begin{table}[t!]
	\begin{center}
		\caption{Accuracy of the variants of IMG module with background class (BG), weighted GAP (wGAP), and feature smoothing (FS), based on the ResNet50 backbone without REG}
		\label{tab:cam_comp}
		\vspace{0.1cm}
		\scalebox{0.85}{
			\begin{tabular}{cccccc}
				\toprule
				& BG & BG + wGAP & BG + FS  & wGAP + FS & All \\ \hline\hline
				 mAP$_{0.5}$&28.8&30.0&31.8&27.4  &  \bf{33.7}  \\
				\bottomrule
			\end{tabular}
		}
	\end{center}
	\vspace{-0.3cm}
\end{table}

\subsubsection{IMG module}
We further investigate the components in the IMG module and summarize the results in Table~\ref{tab:cam_comp}.
All results are from the experiments without bounding box regression to demonstrate the impact of individual components clearly.
All the three tested components make substantial contribution for performance improvement.
The background class activation map models background likelihood within a bounding box explicitly and facilitates the comparison with foreground counterparts.
The feature smoothing regularizes excessively discriminative activations in the inputs to CAM module while the weighted GAP pays more attention to the proper region for segmentation.

\subsubsection{Comparison to a Simple Algorithm Combination}
To demonstrate the benefit of our unified framework, we compare the proposed algorithm with a straightforward combination of weakly supervised object detection and semantic segmentation methods.
Table~\ref{tab:end2end} presents the result from a combination of weakly supervised object detection algorithm, OICR~\cite{tang2017multiple}, and a weakly supervised semantic segmentation algorithm, AffinityNet~\cite{ahn2018learning}.
Note that both OICR and AffinityNet are competitive approaches in their target tasks.
We train the two models independently and combine their results by providing a segmentation label map using AffinityNet for each detection result obtained from OICR.
The proposed algorithm based on a unified end-to-end training outperforms the simple combination of two separate modules even without post-processing.

\begin{table}[t]
\begin{center}
\caption{Comparison our model with a combination of OICR and AffinityNet on PASCAL VOC 2012 segmentation \textit{val} set} 
\label{tab:end2end}
		\vspace{0.2cm}
\scalebox{0.85}{
\begin{tabular}{cccc}
   \toprule
	Model & $\begin{array}{c} \text{OICR} \\ \text{+ AffinityNet} \end{array}$  & $\begin{array}{c} \text{OICR (ResNet50)} \\ \text{+ AffinityNet} \end{array}$ & Ours\\ \hline\hline
  	mAP$_{0.5}$ & 27.3 & 33.3 & \bf{35.9}  \\
   \bottomrule
\end{tabular}
}
\end{center}
\vspace{-0.2cm}
\end{table}

{\setlength{\tabcolsep}{0.5em}
\begin{table}[t]
\begin{center}
\caption{Comparison of class-agnostic regressor and class-specific regressor into our algorithm in terms of detection performance. The evaluation is performed on PASCAL VOC 2012 segmentation \textit{val} set for mAP and \textit{trainaug} set for CorLoc.} 
\label{tab:reg}
\vspace{0.2cm}
\scalebox{0.85}{
\setlength\tabcolsep{15pt}
\begin{tabular}{cccc}
   \toprule
	Model  & mAP  & CorLoc\\ \hline\hline
  	 Ours w/o REG & 49.2 & 66.8  \\
	   Ours (class-specific) &50.4  & 68.4 \\
	   Ours (class-agnostic) & \textbf{53.2} & \textbf{70.1} \\
   \bottomrule
\end{tabular}
}
\end{center}
\vspace{-0.2cm}
\end{table}
}

\begin{figure*}[t]
\begin{center}
   \includegraphics[width=0.85\linewidth]{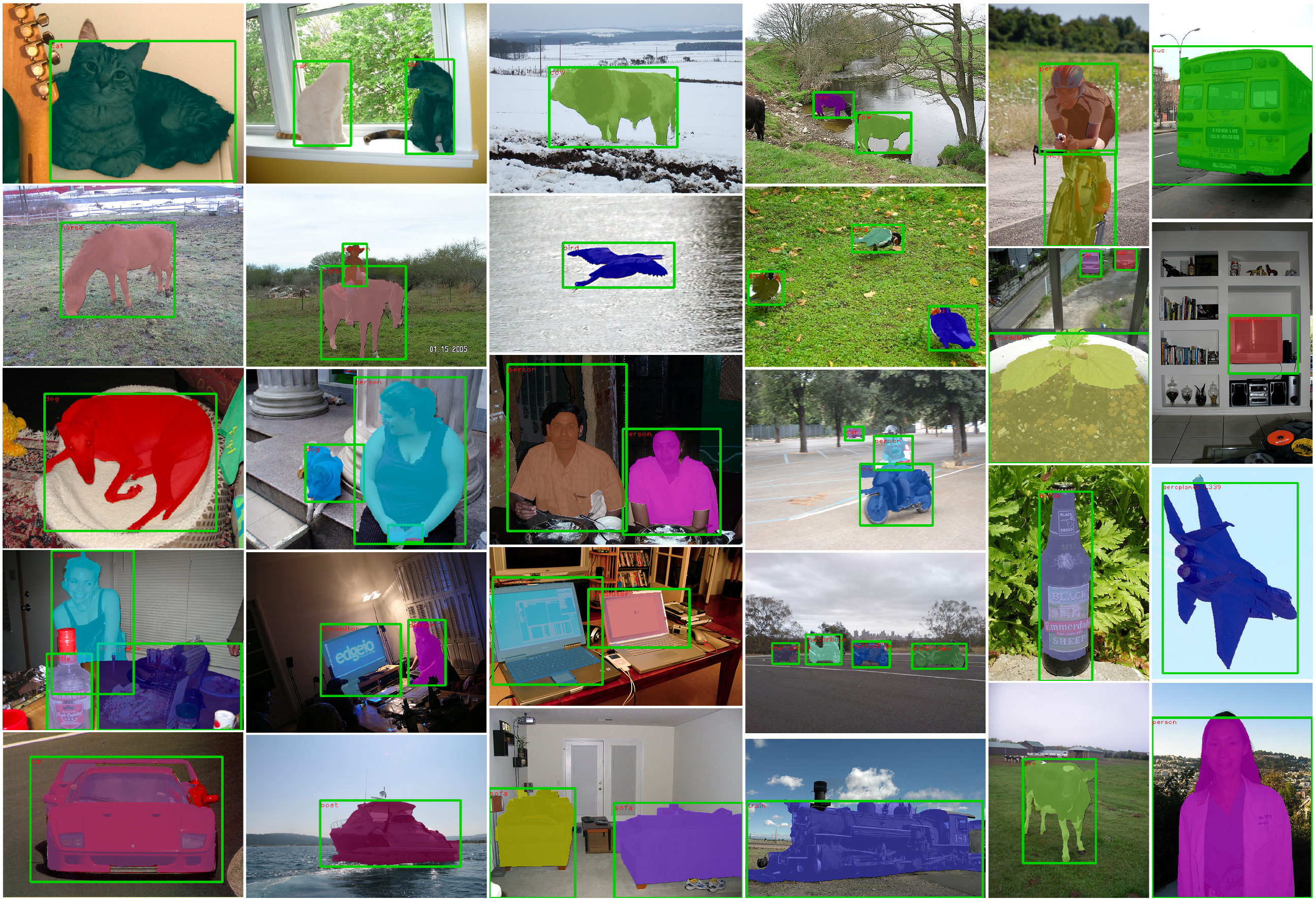}
\end{center}
\vspace{-0.3cm}
   \caption{Instance segmentation results on PASCAL VOC 2012 segmentation \textit{val} set. Green rectangle is a detected object bounding box.}
\label{fig:seg}
\vspace{-0.1cm}
\end{figure*}

\begin{figure*}[t!] 
\begin{center}
	\includegraphics[width=0.85\linewidth]{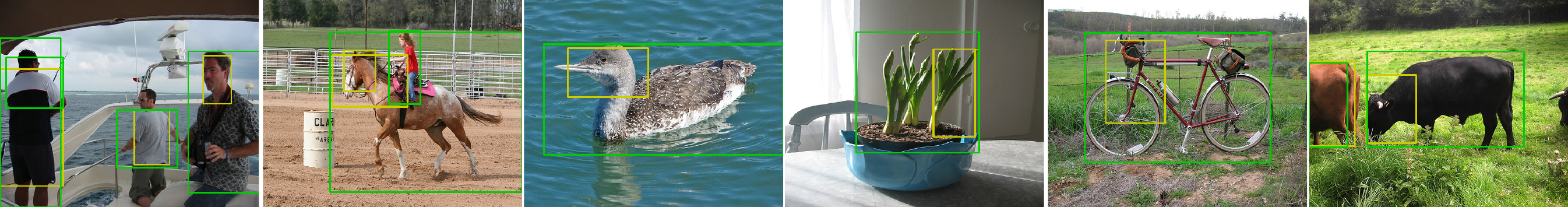}
\end{center}
\vspace{-0.2cm}
   \caption{Qualitative results of detection on PASCAL VOC 2012 segmentation \textit{val} set. Green rectangle is generated by our model and yellow one indicates the output of detector-only model (OICR~\cite{tang2017multiple} based on ResNet50).
   }
\label{fig:det}
\vspace{-0.3cm}
\end{figure*}

\subsubsection{Comparison to Class-Specific Box Regressor}

We compare the results from the class-agnostic and class-specific bounding box regressors in terms of mAP and CorLoc.
Table~\ref{tab:reg} presents that bounding box regressors turn out to be learned effectively despite incomplete supervision.
It further shows that the class-agnostic bounding box regressor clearly outperforms the class-specific version.
We believe that this is partly because sharing a regressor over all classes reduces the bias observed in individual classes and regularizes overly discriminative representations.

\subsection{Qualitative Results}
Figure~\ref{fig:seg} shows instance segmentation results from our model after post-processing and identified bounding boxes on PASCAL VOC 2012 segmentation \textit{val} set.
Refer to our supplementary material for more details.
Our model successfully segments whole regions of objects and discriminates each object in a same class within an input image via predicted object proposals.
Figure~\ref{fig:det} compares detection results from our full model and a detector-only model, OICR with the ResNet50 backbone network, on the same dataset.
Our model is more robust to localize a whole object since the features are better regularized by joint learning of Object Detection, IMG, and Instance Segmentation modules.

%% file: sections/5_conclusion.tex
\section{Conclusion}\label{sec:conclusion}	

We presented a unified end-to-end deep neural network for weakly supervised instance segmentation via community learning.
Our framework trains three subnetworks jointly with a shared feature extractor, which performs object detection with bounding box regression, instance mask generation, and instance segmentation.
These components interact with each other closely and form a positive feedback loop with cross-regularization for improving quality of individual tasks.
Our class-agnostic bounding box regressor successfully regularizes object detectors even with weak supervisions only while the post-processing based on MCG mask proposals improves accuracy substantially.

The proposed algorithm outperforms the previous state-of-the-art weakly supervised instance segmentation methods and the weakly supervised object detection baseline on PASCAL VOC 2012 with a simple segmentation module.
Since our framework does not rely on particular network architectures for object detection and instance segmentation modules, using better detector or segmentation network would improve the performance of our framework.

%% file: sections/6_supp.tex

\setcounter{section}{0}
\renewcommand{\thesection}{\Alph{section}}

\begin{figure*}[t!]
	\begin{center}
		\includegraphics[width=0.95\linewidth]{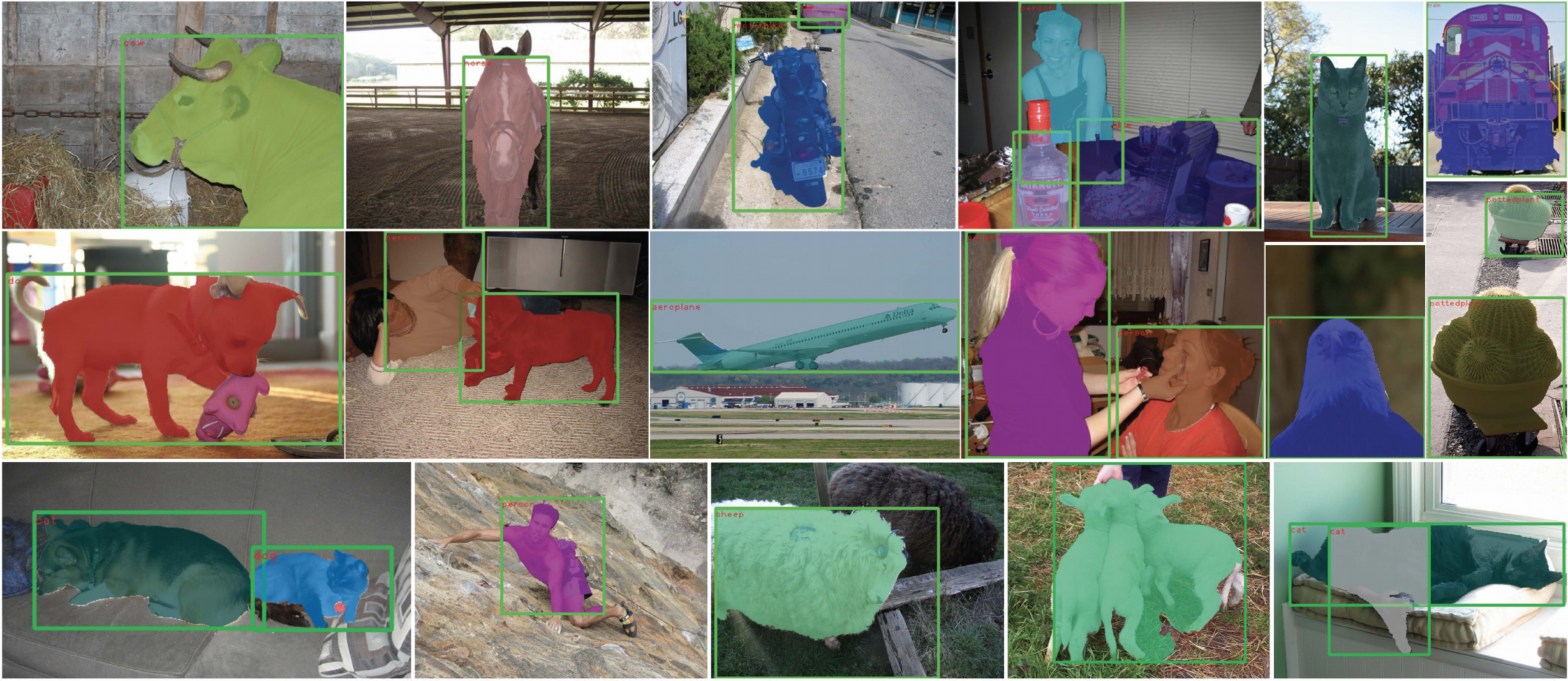}
	\end{center}
	\caption{Qualitative results of instance segmentation on PASCAL VOC 2012 segmentation \textit{val} set.
	Results in the first two rows are the success cases and those in the last row are failure cases}
	\label{fig:seg_qual}
\end{figure*}

\section*{Appendix}

\section{Details of Our Framework}
This section discusses more details regarding our feature extractor, object detection and instance mask generation modules, which are described in our main paper.

\subsection{Feature Extractor}
\label{sub:backbone}
We use ResNet50~\cite{he2016deep} as a backbone network, which is pretrained on ImageNet.
For object detection, one SPP layer is attached after \textsf{res4}, followed by \textsf{res5}. 
The output of the last residual block is shared with IMG and segmentation modules through upsampling.
The IMG module employs multiple level of $28\times 28$ features from outputs of SPP layers attached to  \textsf{res3} and \textsf{res4}, and upsampled \textsf{res5} output.
These features are given to the weighted GAP and the classification layers following one convolution layer for each level of the CAM subnetwork.
For instance segmentation, the upsampled output of \textsf{res5} is used.
On our implementation, batch normalization is replaced to group normalization~\cite{wu2018group} due to the small batch size.

\subsection{Object Detection Module}
Object detection module is composed of detector and regressor parts.
Note that any weakly supervised object detection algorithm can be used as the detector in the proposed framework.

\subsubsection{Detector}%
We adopt OICR~\cite{tang2017multiple} for the detector.
OICR is one of the most commonly used algorithm for weakly supervised object detection relying on multiple instance learning~\cite{tang2018pcl,tang2017multiple,zhang2018w2f}.
The model has two parts, multiple instance detection network (MIDN) and refinement layers.%

\subsubsection{MIDN}
MIDN is based on the Weakly Supervised Deep Detection Network (WSDDN)~\cite{bilen2016weakly}, which has two parallel fully connected layers for classification and detection, respectively and they are followed by two separate softmax layers.
For classification, the softmax layer is given by
\begin{equation}
	[\sigma_{\text{cls}}(\mathbf{x}^c)]_{ij} = \frac{e^{x_{ij}^c}}{\sum_{k=1}^C e^{x^c_{kj}}},
\end{equation}
where $x^c_{ij}$ denotes the classification score for the $i^\text{th}$ class of the $j^\text{th}$ proposal and $C$ denotes the number of classes.
On the other hand, the softmax layer for detection branch is given by
\begin{equation}
	[\sigma_{\text{det}}(\mathbf{x}^d)]_{ij} = \frac{e^{x_{ij}^d}}{\sum_{k=1}^{|{R}|} e^{x^d_{ik}}},
\end{equation}
where $x^d_{ij}$ denotes the detection score for the $i^\text{th}$ class of the $j^\text{th}$ proposal and $|{R}|$ is the number of proposals.

The final score, $\mathbf{z} \in \mathbb{R}^{C \times |{R}|}$ is defined as 
\begin{equation}
	\mathbf{z} =  \sigma_{\text{cls}}(\mathbf{x}^c) \odot \sigma_{\text{det}}(\mathbf{x}^d), 
\end{equation}
where $\odot$ is the Hadamard product.
The image-level classification score $\boldsymbol{\phi}$ is given by the sum of $\mathbf{z}$ over all proposals.
By using the image-level score, the loss from MIDN\, $\mathcal{L}_\text{cls}$ is defined as an image-level cross-entropy, which is described in Eq.~3 in our main paper.

\subsubsection{Refinement Layer}
Once MIDN predicts a class of each proposal, a refinement layer revises the labels by leveraging object classification scores from the previous stage.
The refinement layer  finds the proposal with the highest rank in each class, which is considered as a seed.
Each proposal is given a label from the highest overlapping seed if its IoU (Intersection over Union) with the seed is higher than a threshold, 0.5; otherwise, it is labeled as a background class.
The weight of the proposal $w_r$ is given by the class score of the seed.
Hence, the loss of the $k^{\text{th}}$ refinement layer,  $\mathcal{L}_\text{refine}^k$ is defined as a weighted cross-entropy loss as described in Eq.~4 in our main paper.

\subsubsection{Regressor}
For bounding box regression, we attach two fully connected layers after \textsf{res5} which has 2048 channels. 
The final output of our regressor has a dimension of 4 for class-agnostic manner instead of $4C$ where $C$ is the number of classes for traditional class-specific manner.
It means that class-agnostic regressor is shared with all classes.

During training, a proposal and its nearest pseudo-ground-truth proposal pair $( \text{p}, g)$ is converted to a regression offset $t = [t_{x},t_{y},t_{w},t_{h}]$ as follows:
\begin{equation}
\begin{split}
	t_{x} &= (g_{x} -  \text{p}_{x}) /  \text{p}_{w},\\
	 t_{y} &= (g_{y} - \text{p}_{y}) /  \text{p}_{h},  \\
	t_{w} &= \log(g_{w}/ \text{p}_{w}), \\
	 t_{h} &= \log(g_{h}/ \text{p}_{h}),
\end{split}
\label{eq:reg_offset}
\end{equation}
where  $g = [g_{x}, g_{y}, g_{w}, g_{h}]$ is a target pseudo-ground-truth proposal for a proposal, $\text{p} = [ \text{p}_{x},  \text{p}_{y},  \text{p}_{w},  \text{p}_{h}]$. 

\vspace{0.5cm}
\subsection{Instance Mask Generation (IMG) Module}
We use CAM~\cite{zhou2016learning} for instance mask generation module.
It can be substituted by other object localization algorithms based on image-level labels such as Grad-CAM~\cite{selvaraju2017grad} and Grad-CAM++~\cite{chattopadhyay2017grad}.

\subsubsection{Class Activation Map (CAM)}
CAM~\cite{zhou2016learning} highlights areas of discriminative parts of objects over each class and is often used for the pseudo-ground-truth for weakly supervised semantic segmentation. 
CAM is built on a classification task leveraging Global Average Pooling (GAP)~\cite{lin2013network}.
It is applied to the last convolutional layer followed by a fully connected layer and a softmax layer to predict image-level class labels.
For each class c, CAM, $\mathbf{M}_{c}(x,y)$ is defined as follows:
\begin{equation}
 \mathbf{M}_{c}(x,y) = \mathbf{w}_{c}^T\cdot \mathbf{F}(x,y),
\end{equation}
where $\mathbf{F}(x,y)$ is a feature vector from the last convolutional layer with respect to spatial grid $(x,y)$,
and $\mathbf{w}_c$ is a weight vector of fully connected layer.

\section{Time Cost of Post Processing}
Note that our model without post-processing has competitive results compared to existing methods, and our post-processing is finding best matching MCG proposal for each predicted mask.
The computational cost for post-processing is not signficant compared to our main algorithm based on a deep neural network.
Specifically, the inference through our network takes 4 seconds per image (5 multi-scales with flip) on a single TITAN Xp GPU but the post-processing takes $0\sim4$ seconds on a CPU.

\begin{figure}[t]
	\begin{center}
		\includegraphics[width=0.85\linewidth]{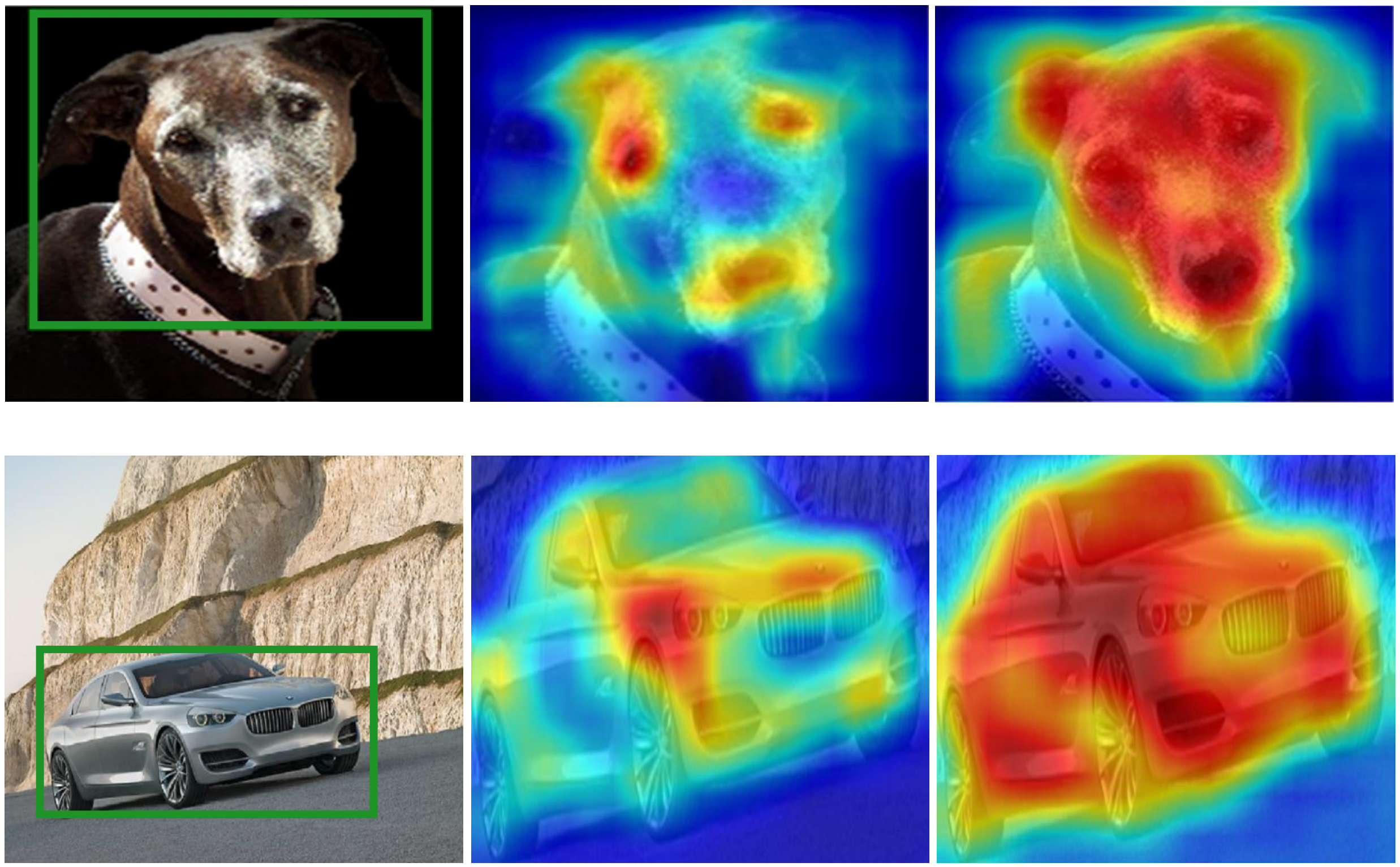}
	\end{center}
		\vspace{-0.2cm}
	\caption{Comparison between the outputs of the conventional CAM network (middle) and one with feature smoothing (right) for two images.}
	\label{fig:non-linear}

\end{figure}

\begin{figure*}[t]
	\begin{center}
		\includegraphics[width=0.95\linewidth]{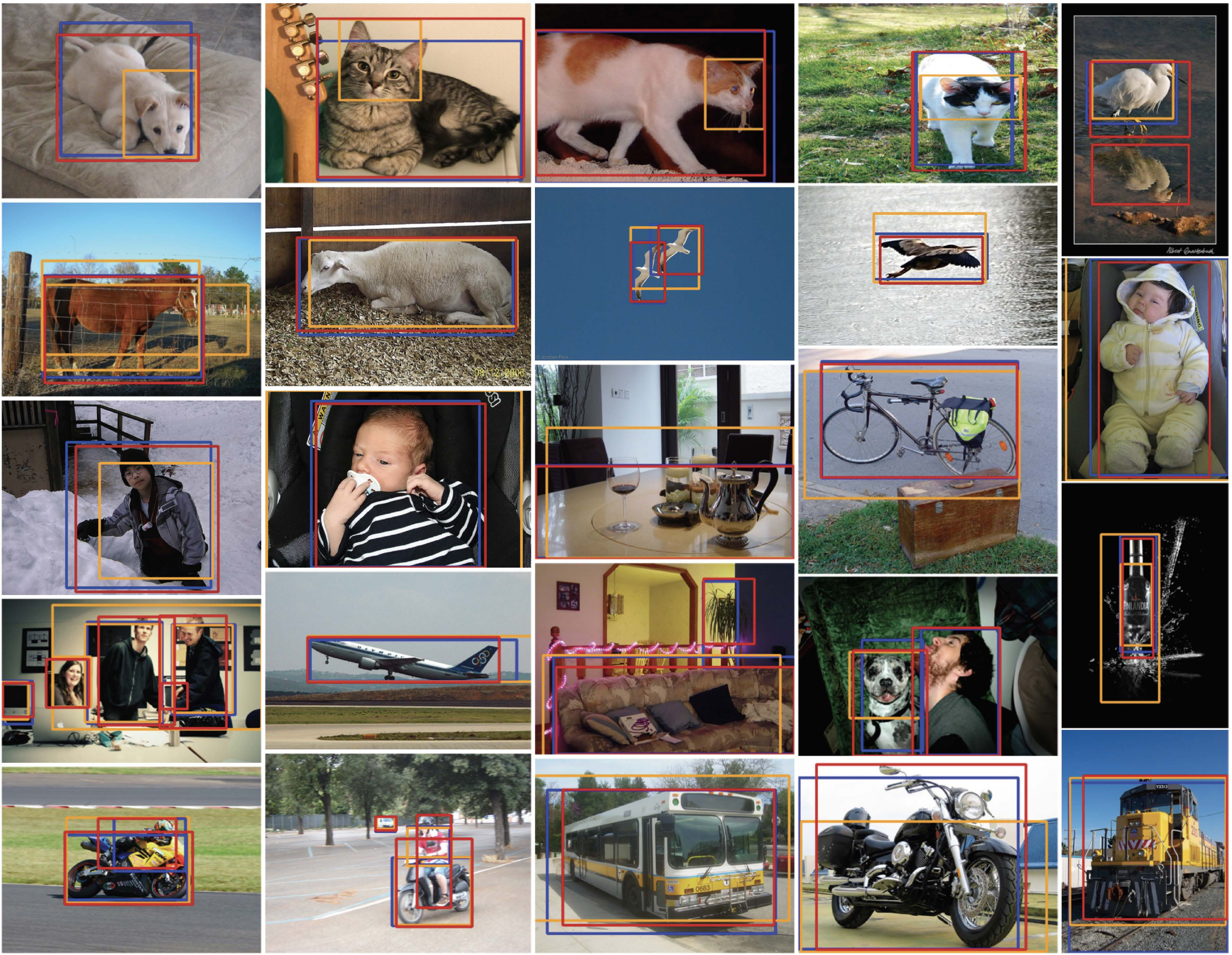}
	\end{center}
	\caption{Qualitative results regarding class-specific and class-agnostic regressors on PASCAL VOC 2012 segmentation \textit{val} set. Red rectangle is a ground-truth, blue rectangle represents the output of our model with class-agnostic regressor and orange one is our model with class-specific regressor.}
	\label{fig:reg_result_2}
\end{figure*}

{\setlength{\tabcolsep}{0.5em}
\begin{table}[t]
\begin{center}
\caption{Accuracy of the various number of CAMs in IMG module based on ResNet50 without REG and IS modules on PASCAL VOC 2012 segmentation \textit{val} set.} 
\label{tab:multi_cams}
\vspace{0.3cm}
\scalebox{1.0}{
\setlength\tabcolsep{5pt}
\begin{tabular}{ccccc}
   \toprule
	The number of CAMs & 1 & 2 & 3 (ours) & 4\\ \hline\hline
  	mAP$_{0.5}$ & 29.7 & 30.6 & \textbf{32.8}  & 31.3  \\
   \bottomrule
\end{tabular}
}
\end{center}
\vspace{-0.2cm}
\end{table}
}

\section{Additional Ablation Study}
\label{sec:abl}	

\subsection{Multiple CAMs}
We present the instance segmentation performance at mAP$_{0.5}$ with respect to the number of CAMs in Table~\ref{tab:multi_cams}.
The results come from our Detector + IMG module  which does not have REG and IS modules and the postprocesing to directly show the effectiveness of multiple CAMs.
The multi-scale representations are helpful to capture whole objects rather than discriminative parts only.

\section{Additional Qualitative Results}
\label{sec:qual}	
\subsection{Instance Segmentation}
Figure~\ref{fig:seg_qual} shows additional instance segmentation results.
Images in the first two rows are success cases and those in the last row are failure cases.
In the failure cases, the model is confused with dog and cat and cannot detect human hands and leg, dark sheep. differentiate adjacent three sheep, and remove false positive.

\subsection{Feature Smoothing}
To penalize CAM focusing excessively on discriminative parts on target objects, we smooth the input features to CAM networks using a non-linear activation function.
As illustrated in Figure~\ref{fig:non-linear}, the function helps produce more spatially regularized activation maps which are more appropriate to enclose entire target objects by segmentation.

\subsection{Bounding Box Regression}
\label{subsubsec:qual_reg}
\begin{figure*}[]
\begin{center}
      \includegraphics[width=0.95\linewidth]{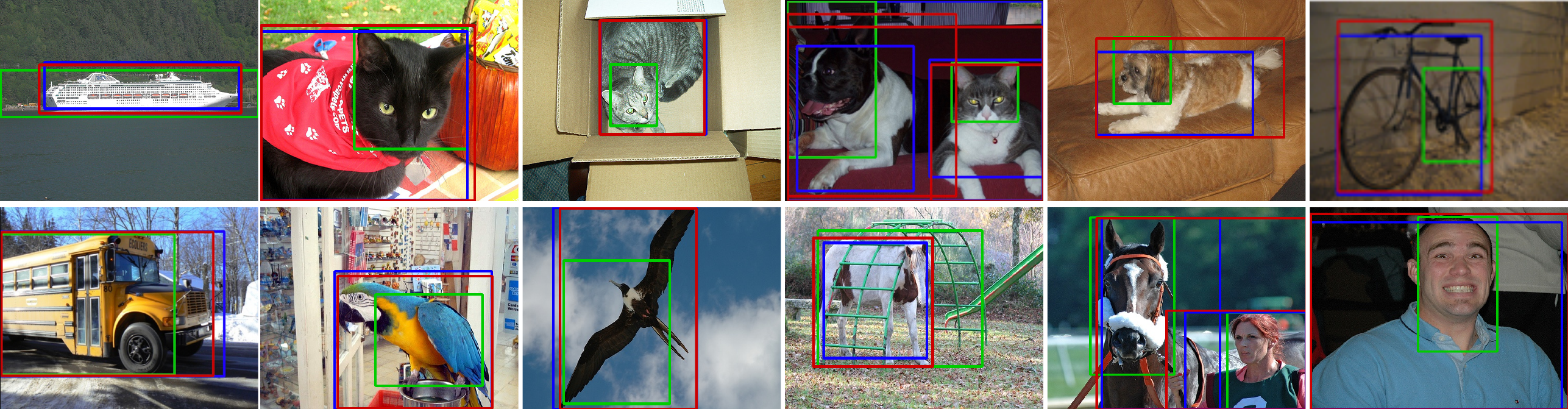}
\end{center}
	\caption{Qualitative results of detection on PASCAL VOC 2012 segmentation \textit{val} set. Red rectangle indicates ground-truth, green rectangle is generated by our model without regressor, and blue one represents the output of our model with class-agnostic regressor.}
\label{fig:reg_supp}
\end{figure*}

We qualitatively compare our model with class-agnostic regressor and with class-specific regressor on Figure~\ref{fig:reg_result_2}.
Our model with class-agnostic regressor achieves better performance than with class-specific regressor.
The difference between two regressors is remarkable on ``cat'' and ``dog'' classes.
With class-agnostic regressor, our model detects their entire bodies while the model with class-specific counterpart still spotlights their discriminative parts, faces.
Figure~\ref{fig:reg_supp} presents the effectiveness of our class-agnostic regressor compared to our model without regressor on PASCAL VOC segmentation \textit{val} set.

